\documentclass[runningheads]{llncs}

\usepackage{eccv}

\usepackage{eccvabbrv}

\usepackage{graphicx}
\usepackage{booktabs}
\usepackage{algorithm}
\usepackage{algorithmic}

\usepackage[accsupp]{axessibility}  

\usepackage{hyperref}

\usepackage{orcidlink}

\begin{document}

\title{Incremental Open-set Domain Adaptation} 


\author{Sayan Rakshit$^*\dag$\inst{1}\orcidlink{0000−0002−0189−7257} \and
Hmrishav Bandyopadhyay$^*\ddag$\inst{2,3} \and
Nibaran Das\inst{3}\and
Biplab Banerjee\inst{1}\orcidlink{0000−0001−8371−8138}}

\authorrunning{S.~Rakshit et al.}

\institute{Indian Institute of Technology Bombay \and
University of Surrey \and
Jadavpur University \\
\email{$\dag$ sayan1by2@gmail.com, $\ddag$ hmrishavbandyopadhyay@gmail.com}\\
}

\maketitle
\makeatletter\def\Hy@Warning#1{}\makeatother
\def\thefootnote{*}\footnotetext{Equal Contribution, $\ddag$ Work done at Jadavpur University}\def\thefootnote{\arabic{footnote}}

\begin{abstract}
  Catastrophic forgetting makes neural network models unstable when learning visual domains consecutively. The neural network model drifts to catastrophic forgetting-induced low performance of previously learnt domains when training with new domains. We illuminate this current neural network model weakness and develop a forgetting-resistant incremental learning strategy. Here, we propose a new unsupervised incremental open-set domain adaptation (IOSDA) issue for image classification. Open-set domain adaptation adds complexity to the incremental domain adaptation issue since each target domain has more classes than the Source domain. In IOSDA, the model learns training with domain streams phase by phase in incremented time. Inference uses test data from all target domains without revealing their identities. We proposed IOSDA-Net, a two-stage learning pipeline, to solve the problem.  The first module replicates prior domains from random noise using a generative framework and creates a pseudo source domain. In the second step, this pseudo source is adapted to the present target domain. We test our model on Office-Home, DomainNet, and UPRN-RSDA, a newly curated optical remote sensing dataset.
\end{abstract}

\section{Introduction}
\label{sec:intro}

\begin{figure}[ht]
    \centering
    \includegraphics[width=0.95\linewidth, height=6cm]{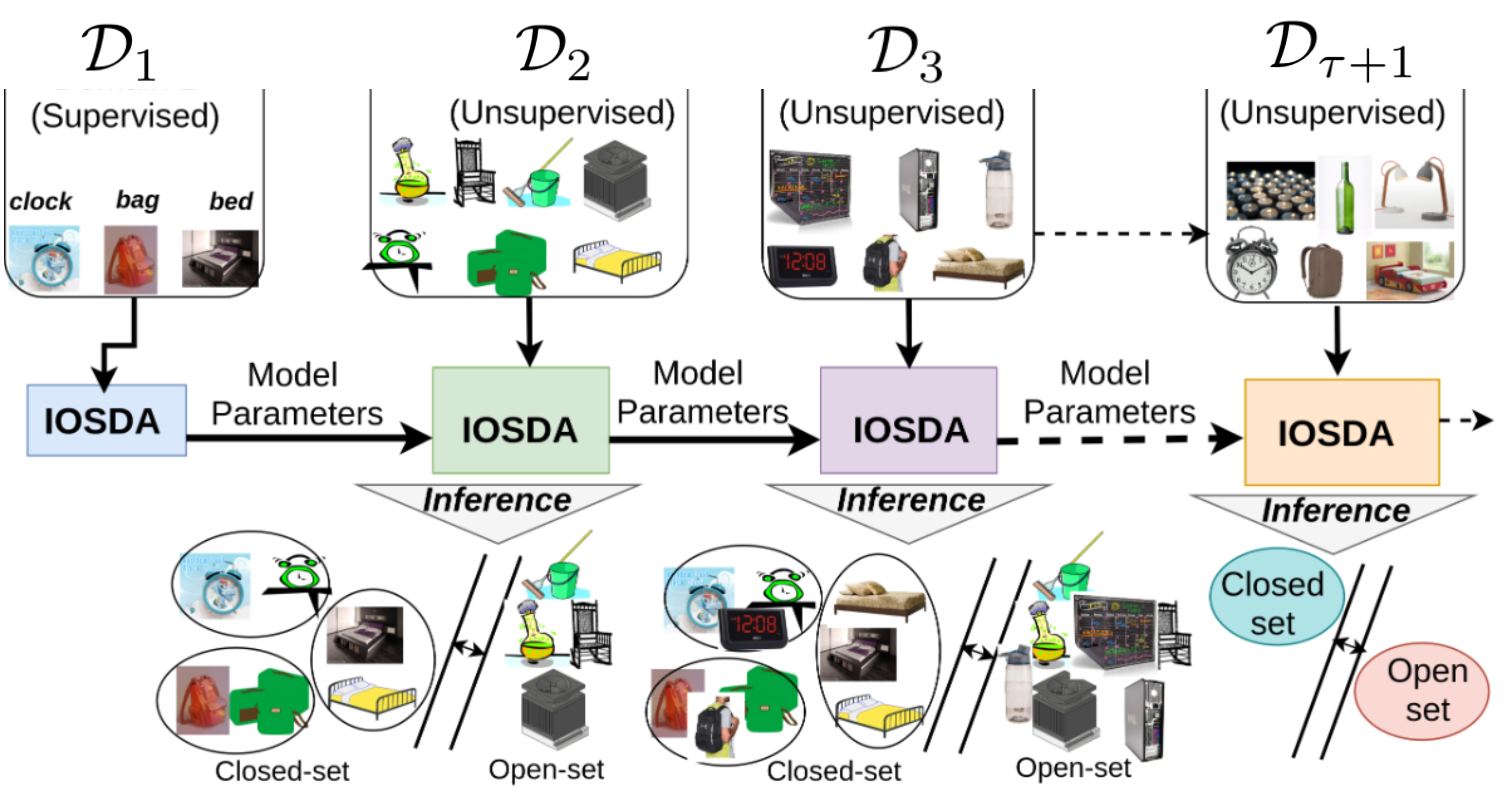}
    \caption{Illustrating the flow of domain data across different time-stamps, IOSDA performs lifelong open-set domain adaptive classification tasks. At each time-stamp, IOSDA encounters a single previously unseen domain and retains model parameters from previous time-stamps. IOSDA classifies closed-set object classes based on their known classes and identifies unseen classes as open-set. During inference, data can originate from any previously encountered domains up to the current time-stamp.}
    \label{fig:pic1}
\end{figure}

Deep learning models have achieved significant advancements in visual inference tasks, primarily due to abundant labeled datasets. However, traditional CNNs excel in stationary data distributions but struggle with non-stationary ones, like those in domain adaptation scenarios, where the training and test data distributions differ. Handling cross-domain challenges, especially in scenarios with previously unseen classes (Open-set recognition), further complicates specific task performance, such as classification in unsupervised target domains. This realism is compounded when domains appear incrementally over time without overlap between timestamps, limiting access to previous domains due to memory or privacy constraints. Combining these challenges (Open-set recognition and Incremental learning) creates a real-world scenario, reducing annotation costs and eliminating the need for source-target domain coexistence in conventional domain adaptation training. This introduces a new problem setup known as Incremental Open-set Domain Adaptation to the computer vision community, integrating domain adaptation, Open-set recognition, and Incremental learning into a unified visual recognition system for efficient inference.

Domain adaptation (DA) involves two visual domains, $\mathcal{S}$ (Source) and $\mathcal{T}$ (Target), with the assumption $\mathcal{P}(\mathcal{S}) \neq \mathcal{P}(\mathcal{T})$. Typically, $\mathcal{S}$ is labeled while $\mathcal{T}$ is unlabeled in unsupervised DA. A common approach in DA is to learn a domain-independent feature space where domains overlap, enabling classifiers trained on $\mathcal{S}$ to classify $\mathcal{T}$ samples.

Unsupervised DA can be closed-set if $\mathcal{S}$ and $\mathcal{T}$ share the same image categories. If $\mathcal{T}$ includes additional classes beyond $\mathcal{S}$, it's termed open-set DA (OSDA), presenting challenges of domain-shift and out-of-distribution samples. Adversarial learning strategies are used to distinguish unknown samples from known classes in OSDA.

Incremental learning involves handling multiple tasks with task-specific data acquired at different time stamps. Class incremental learning is a popular incremental learning variant where non-overlapping class subsets are learned sequentially. Data for current classes are accessible, but previous data may be inaccessible due to security or sensitivity concerns. The model faces catastrophic forgetting, which is a phenomenon where model forgets previous knowledge as it learns new classes. Efforts focus on balancing stability and plasticity in incremental learning, ensuring retention of past knowledge while integrating new information. Recently, domain incremental learning has emerged, adapting a model trained on a labeled source domain $\mathcal{S}$ to sequential  adopt unlabeled target domains $\mathcal{T}_1$ and $\mathcal{T}_2$ over different timestamp. However, existing methods assume closed-set scenarios with fixed classes across all domains and continuous availability of $\mathcal{S}$, limiting effectiveness in scenarios where $\mathcal{S}$ becomes unavailable, and $\mathcal{T}$ domains may include open-set classes.

 In this paper, we introduce the Incremental Open-Set Domain Adaptation (IOSDA) problem, crucial for real-world visual inference systems. IOSDA deals with incremental data acquisition over time, varying data distributions, and frequent introduction of novel classes without annotations. Applications include medical imaging, remote sensing, robotics, and real-time object recognition scenarios. Large language-based foundation \cite{devlin2018bert, radford2021learning} models may poses substantial security and privacy risks due to their training on extensive datasets, potentially exposing sensitive information and vulnerabilities exploitable by malicious actors. These concerns drive our focus on developing solutions for IOSDA using feature replay models, avoiding reliance on foundation models.

In IOSDA, sequence of domains $(\mathcal{D}1, \mathcal{D}2, \ldots, \mathcal{D}{\tau}, \mathcal{D}{\tau+1})$ obtained at distinct time stamps $\tau \in {1, 2, \ldots, \tau, \tau+1}$. $\mathcal{D}1$ is the labeled source domain, while subsequent domains $(\mathcal{D}2, \ldots, \mathcal{D}{\tau}, \mathcal{D}{\tau+1})$ are unlabeled and acquired sequentially. All domains share a set of classes termed as "Known" or "Closed-set." Additionally, each target domain contains extra classes known as "Unknown" or "Open-set," which may be shared or domain-specific. Notably, $\mathcal{D}_1$ becomes unavailable for $\tau > 0$.

The paper's contributions include: (1) Introducing the novel problem IOSDA to the computer vision community. (2) Proposing the MDCGAN domain generation technique to replicate original domain samples. (3) Proposing MEOSDA, an adaptation technique capable of handling multiple domains simultaneously. (4) Use of ensemble technique to enhance performance in multi-domain setups. (5) Curating the UPRN-RSDA optical remote sensing dataset for Incremental Domain Adaptation. (6) Conducting extensive experiments on Office-Home, DomainNet, and UPRN-RSDA datasets.
\section{Related Works}

\textbf{Domain adaptation}: Unsupervised Domain Adaptation (DA) addresses the challenge of domain shift between training and testing data distributions in machine learning. Numerous DA methods (\cite{31, 32, 33, 20, 38}) have enhanced performance in computer vision tasks such as classification, segmentation, and object detection. However, these methods assume that both training and testing domains share the same set of object classes, limiting their applicability in real-world scenarios where target domains lack supervision. Open Set Domain Adaptation (OSDA) addresses this limitation by allowing target domains to include classes not present in the source domain. Early approaches like \cite{OSVM} used class-wise probabilities to distinguish open and closed class samples. \cite{osdabp} introduced adversarial learning to simultaneously segregate open and closed class samples while aligning the closed class distributions between source and target domains. Recent methods such as \cite{STA} deploy adversarial learning with weighting schemes to achieve fine-grained domain alignment. \cite{ROT} and \cite{AOD} utilize self-supervision and contrastive mapping, respectively, to handle open-set samples and align semantically similar closed-set classes across domains.\\

\textbf{Incremental learning}: Incremental learning \cite{7} involves continuously training a classifier on sequentially available data while preserving model performance throughout the learning process to avoid catastrophic forgetting. Class incremental learning \cite{12, 22, 25} is a well-studied variant where regularization techniques penalize parameter changes that could lead to forgetting previous tasks. Dynamic modeling approaches \cite{23, 24} increase model capacity over time to adapt to new tasks. Conditional GAN architectures \cite{19, 28, 29} are utilized to generate synthetic samples, with methods like \cite{19} and \cite{26, 17} employing replay-based techniques using real and synthetic samples to mitigate catastrophic forgetting issues from previous tasks.  \\

\noindent \textbf{Incremental domain adaptation}: Computer vision literature offers advanced DA algorithms. However, Incremental Domain Adaptation (IDA) is understudied. IDA aims to update the parameters of the DA model (trained by adapting $\mathcal{S}$ and $\mathcal{T}_1$) by adapting a new domain $\mathcal{T}_2$ with the previous domains $(\mathcal{S}, \mathcal{T}_1)$. Several methods \cite{45, 11} assume a marginal domain shift, making them suitable for a dynamic context. These methods require prior knowledge of graph-based structure inter-relationships or domain information \cite{40, 44}. A strategy based on memory replay \cite{45, 11, 41} used regularisation or distillation to address catastrophic forgetting.
These approaches do not entirely data-free IDA setup; they need samples from prior domains.
Recent literature \cite{volpi2021continual, FRIDA , taufique2023continual} addresses the issue of domain adaptation, assuming all domains include the same class objects, which is challenging in unsupervised domains. The continual domain adaptation problem is addressed by \cite{volpi2021continual} using meta learning and \cite{FRIDA} using generated feature replay.

\section{Problem Formulation and Methodology}

\noindent \textbf{Problem definition}: We assume the $\tau+1$ visual domains $( \mathcal{D}_1, \mathcal{D}_2, \cdots, \mathcal{D}_\tau, \mathcal{D}_{\tau+1} )$ where samples of $\mathcal{D}_{1}:( \{\textbf{x}^1_i, y^1_i\}_{i=1}^{|\mathcal{D}_1|})$ contains class label information and samples from all other domains $\mathcal{D}_{ m}:(\{\textbf{x}_{i}^{m}\}_{i=1}^{|\mathcal{D}_{\tau}|}), \forall m \in [2,3, ... , \tau+1] $ are unlabeled. All pairs of domains follow non-identical data distribution:$P(\mathcal{D}_m) \neq P(\mathcal{D}_n)$, $1 \leq m, n \leq \tau+1, m\neq n$. 

Our setup consider unsupervised open-set domain adaptation, where only the source domain has label information and target domains may contain a set of open-set and closed-set classes from the source domain without prior knowledge of class labels.  Also, only a domain $\mathcal{D}_{\tau} $ is available at any given timestamp ($\tau$), hence no original source domain is available for adopting new Target domains. We aim to unsupervisedly separate open-set and closed-set classes at the Target domain and classify them into $K+1$ categories, where the first $K$ classes as closed-set and the $K+1^{th}$ as open-set classes. Without access to past domain samples, model adapts to new target domains and maintains performance on all the past domains until $\tau$.\\

\begin{figure}[ht]
    \centering
    \includegraphics[width=0.8\linewidth, height=6.5 cm]{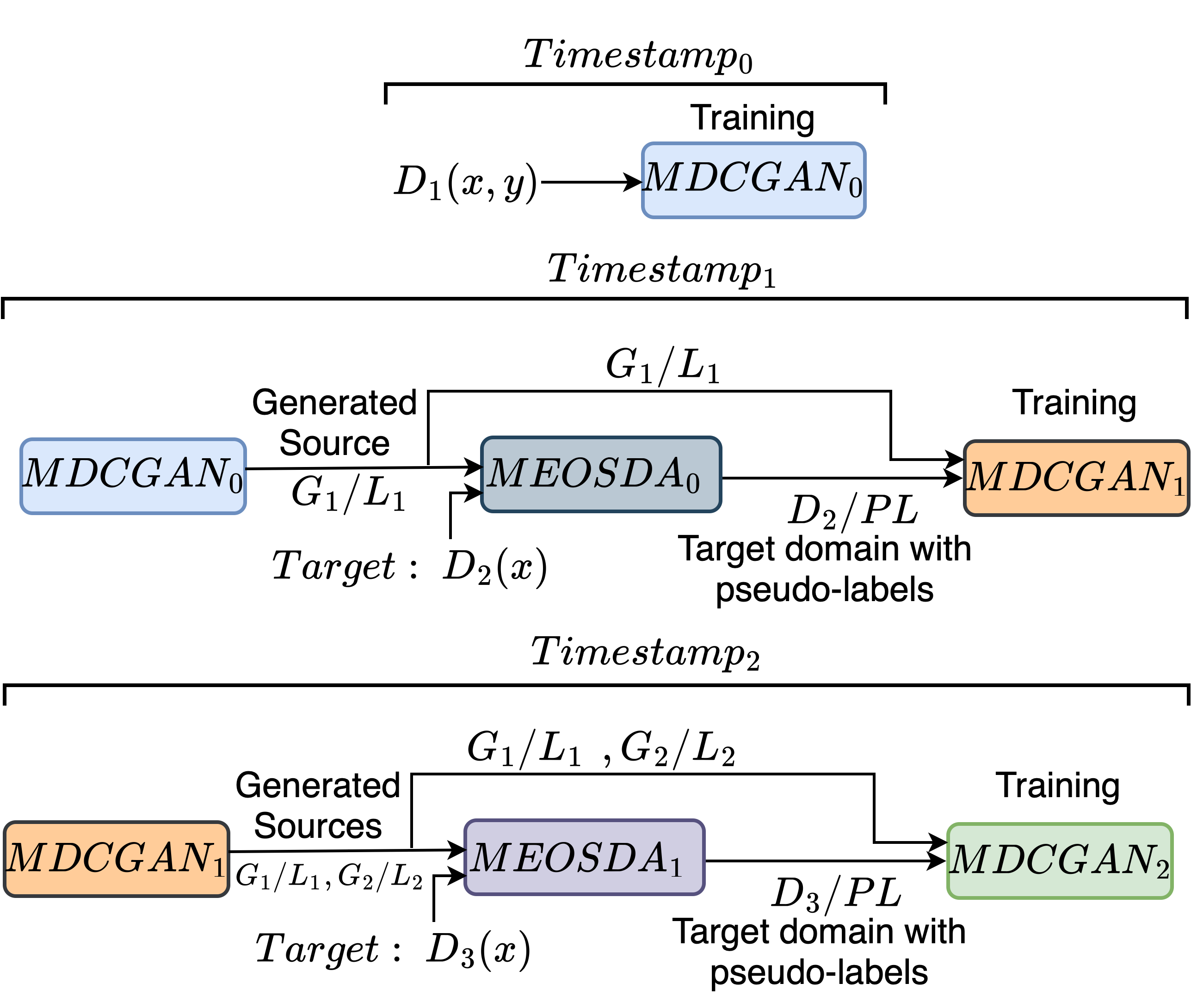}
    \caption{The diagram above shows the entire process flow. One domain $D_1(x,y)$ is available at the initial timestamp. The $MDCGAN_0$ is trained with the original samples of $D_1(x,y)$ to preserve this domain. Next timestamp, a new target domain appears, but the original source domain $D_1(x,y)$ is not available. Instead, $MDCGAN_0$ generates source samples and labels ($G_1/L_1$). At that moment, $MEOSDA_0$ does domain adaptation training using the pseudo source ($G_1/L_1$) as Source domain and original target domain samples ($D_2(x)$). After $MEOSDA_0$ training, target domain samples ($D_2(x)$) were pseudo-labeled. A new $MDCGAN_1$ was trained with generated source and Target domain pseudo labels before this timestamp disappeared. At the next time stamp ($Timestamp_2$), $MDCGAN_1$ generates sources $G_1/L_1$, $G_2/L_2$, which depict as multiple sources ($D_1(x)$, $D_2(x)$ respectively) , and the newly arrived target domain ($D_3(x)$) goes through $MEOSDA_1$ to perform multi-source domain adaptation and then trains a new $MDCGAN_2$ with the generated sources and the target domain with pseudo labels. No raw samples of prior domains are needed; just the $MDCGAN$ learned in the immediate previous timestamp is saved.  }
    \label{fig:pic2}
\end{figure}

\noindent \textbf{Overall Working Principle}: Our IOSDA-Net model has a two-stage learning pipeline (Fig:\ref{fig:pic2}). The initial stage includes a Multi-Domain and Class-Guided Generative Adversary Network ($MDCGAN$) (details in next section). Second stage does incremental Open Set Domain Adaptation using a novel Multi-output Ensembeled IOSDA model ($MEOSDA$) (details in next section). To produce synthetic source data from random noise, $MDCGAN$ uses domain and class categories to create domain specific class-wise samples. $MEOSDA$ performs multi-source openset domain adaptation without original source data samples in the second step.

\begin{algorithm}[H]
\caption{Working principle}
\begin{algorithmic}[1]
\label{alg:the_alg}
\renewcommand{\algorithmicrequire}{\textbf{Input:}}
\REQUIRE $\mathcal{D}_1$,$\mathcal{D}_2$,$\cdots$,$\mathcal{D}_\tau$, $\mathcal{D}_{\tau+1}$ and threshold $Th$
\renewcommand{\algorithmicrequire}{\textbf{Output:}}
\REQUIRE  The trained $MDCGAN_{\tau}$ and $MEOSDA_{\tau-1}$ at each $\tau \in \{1,2, 3, \cdots, \tau\}$
\IF{$\tau=1$}
\STATE Train $MDCGAN_1$ on $\mathcal{D}_1$ 
\ELSIF{$\tau = 2$}
\STATE Generate $\mathcal{D}^{g_1}_1$ using $MDCGAN_1$
\STATE Train $MEOSDA_1$ with Source as $\mathcal{D}^{g_1}_1$ and Target as $\mathcal{D}_2$

\STATE Apply pseudo-labeling with threshold $Th$ to obtain $\hat{\mathcal{D}}_2$ from $\mathcal{D}_2$
\STATE Train $MDCGAN_2$ using  $\mathcal{D}^{g_1}_1 \cup \hat{\mathcal{D}}_2$
 as the real data
\ELSIF{$\tau \geq 2$}
\STATE Generate $G_1 \cup G_2 \cup \cdots \cup G_{\tau}$ using $MDCGAN_{\tau-1}$
\STATE Train $MEOSDA_{\tau-1}$ with $Source= \mathcal{D}^{g_\tau}_1 \cup \mathcal{D}^{g_\tau}_1 \cup \cdots \cup \mathcal{D}^{g_\tau}_\tau$ and $\mathcal{D}_{\tau+1}$ as $Target$ 
\STATE Obtain $\hat{\mathcal{D}}_{\tau+1}$ using pseudo-labeling with $Th$
\STATE Train $MDCGAN_{\tau}$ by considering $\mathcal{D}^{g_\tau}_1 \cup \mathcal{D}^{g_\tau}_1 \cup \cdots \cup \mathcal{D}^{g_\tau}_\tau \cup \hat{\mathcal{D}}_{\tau+1}$ as the real data 
\ENDIF
\end{algorithmic}
\end{algorithm}

\subsection{Multi-Domain and Class guided Generative Adversarial Network (MDCGAN):} 
\begin{figure}[ht]
    \centering
    \includegraphics[width=0.95\linewidth, height=5cm]{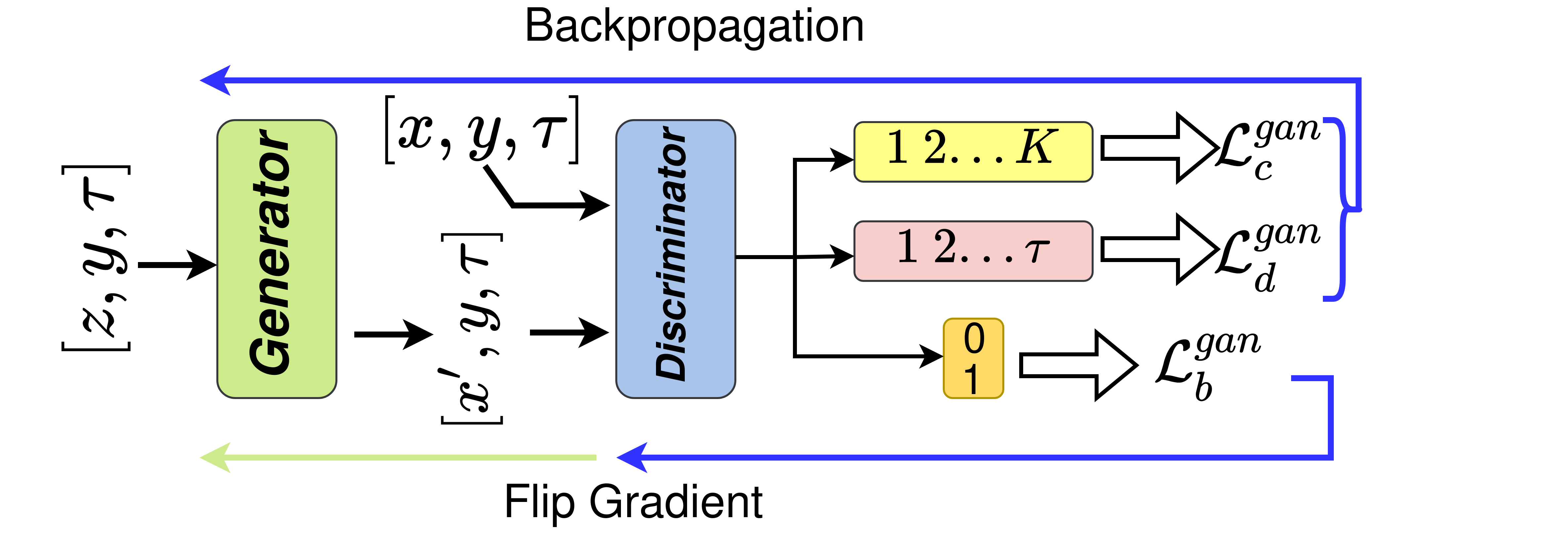}
    \caption{Diagram of Multi-Domain and Class-guided Generative Adversarial Network. Novel MDCGAN has a generator, discriminator, and three output branches (label classifier, domain classifier, and domain discriminator). The discriminator uses the generator's output and the original samples' class and domain labels, while the generator uses random noise. For training, domain discriminator (real/fake) utilises adversarial loss and label, and domain classifier uses cross entropy loss.}
    \label{fig:gan}
\end{figure}

We presented a Multi-Domain and Class-guided Generative Adversarial Network (MDCGAN) (Fig-\ref{fig:gan}) to imitate domains incrementally to maintain samples of absent domains.

\noindent \textbf{MDCGAN-model architecture and overview:}
MDCGAN compose of a feature generator $\mathcal{F}^{gan}(;,\theta_{\mathcal{F}}^{gan})$, one binary classifier $\mathcal{B}^{gan}(;,\theta_{\mathcal{B}}^{gan})$ which discriminate between real domain and fake domain, one auxiliary domains classifier $\mathcal{D}^{gan}(;,\theta_{\mathcal{D}}^{gan})$, and one auxiliary class classifier $\mathcal{C}^{gan}(;,\theta_{\mathcal{C}}^{gan})$. $\mathcal{B}^{gan}$, $\mathcal{D}^{gan}$ and $\mathcal{C}^{gan}$ shares the common multi-layer perceptron network $\mathcal{M}^{gan}$ as a discriminator. Our MDCGAN is based on DGAC-GAN \cite{FRIDA}. Domain separation is crucial for conserving domain-specific information and creating domain-specific samples, however DGAC-GAN doesn't consider it. We attempted to address the shortcomings of DGAC-GAN by adding domain classifier $\mathcal{D}^{gan}$ and other classifiers on top of $\mathcal{M}^{gan}$. 

The feature generator $\mathcal{F}^{gan}$ feed by a concatenated vector of $[z,y,\tau]$, where noise sample $\textbf{z} \in \mathcal{N}(0, \mathbb{I})$, $y$ represent class labels and $\tau$ represents domain labels. The $z$ is truly a random noise that is not dependable to $y$ and $\tau$ but according to $y$ and $\tau$ the $\mathcal{F}^{gan}$ generate the sample from any $z$.  The output of $\mathcal{F}^{gan}$ is  $\mathcal{F}^{gan}([z,y,\tau])$ = $[x']$. Then  $[x', y, \tau]$ and $[x, y, \tau]$ both feeds to $\mathcal{M}^{gan}$ while $[x', y, \tau]$, $[x, y, \tau]$ consider fake, real samples respectively. The real samples $[x,y,\tau]$ (generated samples in the previous timestamp and/or original samples at the current timestamp) and the fake samples$[x', y, \tau]$ are both conditioned on the same $y$ and $\tau$.\\

\noindent \textbf{MDCGAN-model Losses and training objective:}
The object of the MDCGAN can be written in terms of Log-likelihood is in tree fold.\\ 
i) classify data source either in real or fake (ii) classify each sample in one of the categories of K categories of domain $\mathcal{D}_{1}$. (iii) classify each domain into one of the $\tau \in \{1, 2, ..., \tau \}$ domains.

The loss functions are as follows:

\begin{equation}
\centering
    \mathcal{L}_b^{gan} = \mathbb{E} [\log P(B^{gan}(\textbf{x}) = real | \textbf{x}) + \log P(B^{gan}(\textbf{x}') = fake | \textbf{z})]
\end{equation}

\begin{equation}
\centering
    \mathcal{L}_c^{gan} = \mathbb{E} [\log P({C}^{gan}(\textbf{x}) = y | \textbf{x}) + \log P({C}^{gan}(\textbf{x}') = y | \textbf{z})] 
\end{equation}

\begin{equation}
\centering
    \mathcal{L}_d^{gan} = \mathbb{E} [\log P({D}^{gan}(\textbf{x}) = \tau | \textbf{x}) + \log P({D}^{gan}(\textbf{x}') = \tau | \textbf{z})] 
\end{equation}

Also, we used an $\ell_2$-norm based regularizer

\begin{equation}
    \centering
    \mathcal{R}^{gan} = \mathbb{E}_{x \sim (\tau, y)} ||\mathcal{G}^{gan}([\textbf{z},y,\tau]) - \textbf{x}||_2^2
\end{equation}

We follow adversarial training strategy to train MDCGAN and for this purpose we use reverse gradient strategy. The overall optimization process can be written in two steps, \\
(i) Maximize  $\mathcal{L}_c^{gan} +\mathcal{L}_d^{gan}+\mathcal{L}_b^{gan}$ with respect to parameters ${\theta_{\mathcal{M}}^{gan}, \theta_{\mathcal{C}}^{gan}, \theta_{\mathcal{D}}^{gan}, \theta_{\mathcal{B}}^{gan}}$ \\
(ii) Maximize $\mathcal{L}_c^{gan} +\mathcal{L}_d^{gan}-\mathcal{L}_b^{gan} -\mathcal{R}^{gan}$ with respect to parameter $\theta_{\mathcal{F}}^{gan}$

\subsection{Multi-output Ensembeled Open Set Domain Adaptation}
\begin{figure}[ht]
    \centering
    \includegraphics[width=1\linewidth, height=8 cm]{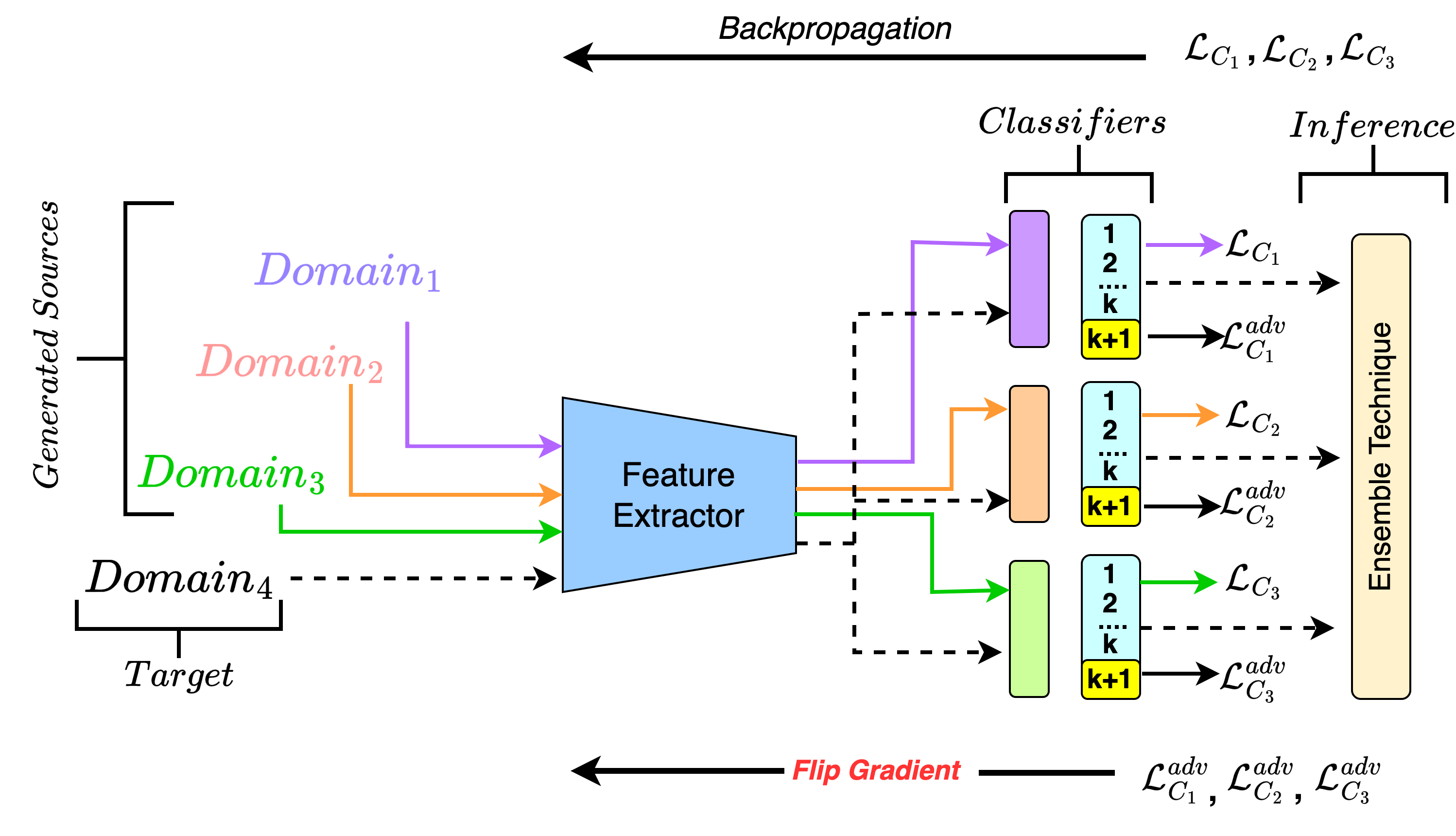}
    \caption{The MEOSDA diagram illustrates the architecture at timestamp 4, featuring $Domain_4$ as the unsupervised target domain and $Domain_1$, $Domain_2$, $Domain_3$ as source domains. In our incremental setup, we employ feature replay to replicate original data for pseudo sources. MEOSDA utilizes a shared feature extractor across all domains, followed by separate classifier branches for each source domain. Each classifier outputs $k+1$ classes, where the first $k$ represent closed classes and the $k+1$th represents open set classes. }
    \label{fig:main}
\end{figure}

\noindent \textbf{MEOSDA-model architecture and overall training procedure:} \\
At $\tau = 1$, only a single domain is available, so there is no scope of domain adaptation. 

When $\tau=2$, we have $\mathcal{S}$: $\mathcal{D}^{g}_1$ (generated samples from MDCGAN that was trained on domain $\mathcal{D}_1$ at the previous timestamp ($\tau = 1$).) and $\mathcal{T}$: $\mathcal{D}_2$. We apply the open-set domain adaptation technique of \cite{osdabp}. At this time stamp $\tau=2$, the model architecture consists of a feature extractor $\mathcal{F}^{osda}(;,\theta_{\mathcal{F}}^{osda})$ and a multi-class classifier $\mathcal{C}_{1}^{osda}(;,\theta_{\mathcal{C}_{1}}^{osda}) $ whose output dimension is K+1 where K+1 th position is representing as open class and rest are closed classes. 

The $\mathcal{C}_{1}^{osda}$ output logit vectors $\{l_1, l_2, ... , l_{K}, l_{K+1}\}$ and on top of it one softmax layer is there to map the logit vector to class probability. The posterior probability for the $k^{th}$ class ($k \in \{1,2,\cdots,K+1\}$) of sample x is mentioned as: $p(y = k|x) = \frac{\exp(l_k)}{\underset{c=1}{\overset{K+1}{\sum}}\exp(l_c)}$.

\begin{equation}
{{\mathcal{L}}_{C_1}}=\mathcal{L}_{CE}(\mathcal{F}^{osda}({\mathcal{C}}_{1}^{osda}(\mathcal{X}_1)), \mathcal{Y}_1),   (\mathcal{X}_1,     \mathcal{Y}_1)\in \mathcal{D}^{g}_1
\label{Eq:2}
\end{equation}

\begin{equation}
\begin{aligned}
   \mathcal{L}_{C_1}^{adv} = \mathbb{E} [- 0.5 \log (p(y_t = K+ 1|\mathcal{C}_{1}^{osda}(\mathcal{F}^{osda}(x_t))\\
    - 0.5 \log(1-p(y_t = K + 1| \mathcal{C}_{1}^{osda}(\mathcal{F}^{osda}(x_t))], \\
    x_t \in \mathcal{D}_2
    \end{aligned}
    \label{Eq35}
\end{equation}

Since $\mathcal{S}$ is rich in class label information, so the classifier $\mathcal{C}_{1}^{osda}$ is trained by a cross-entropy loss ($\mathcal{L}_{CE}$)[\ref{Eq:2}] between the predicted output and ground truth of the labels. To align $\mathcal{S}$ and $\mathcal{T}$ we need to construct a decision boundary between known classes and unknown classes but we are devoid of class label information in Target domain. We utilize a binary cross entropy-based loss [\ref{Eq35}] where $\mathcal{C}_{1}^{osda}$ tries to set a value t (sample of the Target domain) whereas $\mathcal{F}^{osda}$ tries to deceive the $\mathcal{C}_{1}^{osda}$. This mini-max game between $\mathcal{C}_{1}^{osda}$ and $\mathcal{F}^{osda}$ creates a big margin between known and unknown classes in Target domain. We follow \cite{osdabp} and set the t as 0.5.

The overall training objective at $timestamp_2$ ($\tau=2$) is to\\
(i) minimize ($\mathcal{L}_{C_1} + {\mathcal{L}}_{C_1}^{adv})$ with respect to $\theta_{\mathcal{C}_{1}}^{osda}$. \\
(ii)  minimize ($\mathcal{L}_{C_1} - \mathcal{L}_{C_1}^{adv})$ with respect to $\theta_{\mathcal{F}}^{osda}$. We used reverse gradient to accomplish this task. \\

\noindent\textbf{Generalised Objective of MEOSDA at any timestamp: }
So, in general at any given timestamp $\tau$ the $MEOSDA_{\tau-1}$ will be consists of ${\mathcal{C}_{1}}^{osda}, {\mathcal{C}_{2}}^{osda},..., {\mathcal{C}_{\tau}}^{osda}$  classifiers and a shared feature extractor $\mathcal{F}^{osda}$ among all the classifiers. The overall objective function can be formulated as\\
(i)minimize($\sum_{m=1}^{\tau}\mathcal{L}_{C_m}+\mathcal{L}^{adv}_{C_{m}}$)) with respect to ($\theta_{\mathcal{C}_{1}}^{osda}$, ...,  $\theta_{\mathcal{C}_{\tau}}^{osda}$). \\
(ii)minimize($\sum_{m=1}^{\tau}\mathcal{L}_{C_m} - \mathcal{L}^{adv}_{C_{m}}$)) with respect to 
$\theta_{\mathcal{F}}^{osda}$

\subsection{Ensemble during inference:}
In our MEOSDA framework, we incorporate multiple classification heads, each specifically tailored to a domain with identical closed-set classes. During inference, selecting the most confident classification head is pivotal for predicting samples from the target domain, as certain samples from the target domain may bear closer semantic resemblance to specific source domains than others. Predicting samples using classifiers from semantically related domains enhances accuracy on the target domain. Through ensemble techniques, we aim to identify the most confident classification head for each test sample.

The ensemble technique is described pointwise below.
\begin{enumerate}
    \item For each test sample, pass it through each classifier head and record the probability output.
    \item Calculate the absolute difference between the highest probability and the second highest probability for each classifier head. Let these differences be denoted as $DMaxP_1$, $DMaxP_2$, ..., $DMaxP_\tau$ for classifiers $classifier_1$ , $classifier_2$, ..., $classifier_\tau$, respectively.
    \item Similarly, calculate the absolute difference between the lowest probability and the second lowest probability for each classifier head. Let these differences be denoted as $DMinP_1$, $DMinP_2$, ..., $DMinP_\tau$ for classifiers $classifier_1$ , $classifier_2$, ..., $classifier_\tau$, respectively.
    \item If the maximum value among $DMaxP_1$, $DMaxP_2$, ..., $DMaxP_\tau$ and the minimum value among $DMinP_1$, $DMinP_2$, ..., $DMinP_\tau$ both come from the same classifier, consider only the probability output of that classifier for predicting the sample.
    \item If the maximum value among $DMaxP_1$, $DMaxP_2$, ..., $DMaxP_\tau$ and the minimum value among $DMinP_1$, $DMinP_2$, ..., $DMinP_\tau$ come from different classifiers, select the classifier where $DMaxP_i$, $i\in \{1, 2, .. , \tau\}$ is maximized for predicting the sample.
    
    \end{enumerate}

\section{Experiments}
\subsection{Data Set:} Besides Office-Home and Domain-Net, we introduce a new Open Set Domain Adaptation (OSDA) dataset using publicly available remote sensing images for land use and land cover categorization, termed UPRN-RSDA. This dataset includes UCM, PatternNet, RSI-CB256, and NWPU-45 datasets as domains. We will release it soon. Domains were randomly selected from each dataset based on distinct timestamps ($\tau$). Samples in each domain are categorised into known and unknown classes. $\mathcal{D}_1$ exclusively contains known classes with annotated labels. Domains $\mathcal{D}{\tau}, (\tau > 1)$ contain both known and unknown classes, with some unique unknown classes observed only within specific domains. Label information is absent for domains $\mathcal{D}_{\tau}, (\tau > 1)$.\\

\noindent\textbf{Office-Home:} This dataset consists of four domains  and the domains $\mathcal{D}_1$: RealWorld(Rw), $\mathcal{D}_2$: Product(Pr), $\mathcal{D}_3$: Clipart(Cl) and $\mathcal{D}_4$: Artistic(Ar) are chosen respectively at $\tau_0$, $\tau_1$, $\tau_2$, $\tau_3$. There are 65 classes in each domains and in total around 15,500 images are there. We consider $(0-40)$ classes (alphabetically) as known class and rest of the classes as unknown class. The domains $\mathcal{D}_1$, $\mathcal{D}_{2}$, $\mathcal{D}_{3}$ and $\mathcal{D}_{4}$ contains samples of classes ranging from $(0-40), (0-54), (0-40, 45-59)$ and  $(0-40, 45-54, 60-64)$ respectively. The dataset contains classes numbered (alphabetically) from 41 to 64, which are considered open-set classes. Among these classes, 41 to 44, 55 to 59, and 60 to 64 are specific to domains $\mathcal{D}_{2}$, $\mathcal{D}_{3}$, and $\mathcal{D}_{4}$ respectively. This indicates that the open-set class samples differ across the target domains.\

\noindent\textbf{DomainNet:} The dataset DomainNet is a large scale dataset with six visual domains and contains about 0.6 million images across all the domains with 345 classes in each domain. The domains $\mathcal{D}_1$: Real(R), $\mathcal{D}_2$: Painting(P), $\mathcal{D}_3$: Clipart(C) and $\mathcal{D}_4$: Sketch(S), $\mathcal{D}_5$:Quickdraw(Q), $\mathcal{D}_6$:Infograph(I)  are chosen respectively at $\tau_0$, $\tau_1$, $\tau_2$, $\tau_3$, $\tau_4$, $\tau_5$. We consider $(0-239)$ classes (alphabetically) as known class and rest of the classes as unknown class. We split the dataset in such a way  that the domains $\mathcal{D}_1$, $\mathcal{D}_{2}$, $\mathcal{D}_{3}$ , $\mathcal{D}_{4}$ , $\mathcal{D}_{5}$ ,$\mathcal{D}_{6}$ contains samples of classes ranging from $(0-239), (0-284), (0-239, 255-299)$ , $(0-239, 255-284, 300-314)$ , $(0-239, 255-284, 315-330)$ and $(0-239, 255-284, 330-344)$ respectively.  Similar to office-Home dataset, open-set class samples differ across the target domains.\\

\noindent\textbf{UPRN-RSDA:} $\mathcal{D}_1$: UC Merced Land Use Dataset (UCM), $\mathcal{D}_2$: PatternNet(PNet), $\mathcal{D}_3$: RSI-CB256(RSI-CB) and $\mathcal{D}_4$: NWPU-RESISC45(NWPU) are chosen respectively at $\tau_0$, $\tau_1$, $\tau_2$, $\tau_3$. In UCM total 21 classes are there and each class contains 100 images. PNet contains 38 classes and each class contains 800 images. RSI-CB contains 35 classes and total 24000 images are there. NWPU contains 45 classes and per class image are 700. The spatial resolution of the images in UCM, PNet, RSI-CB and NWPU are 0.3, (0.062-4.693), (0.3~3) and (~30 to 0.2) respectively and the image size is 256*256 at each dataset. We find out 7 classes (airplane, beach, forest, freeway, intersection, parking lot, river) that are common among all these four dataset are consider to be known classes. For our setup we consider only these 7 classes from UCM with label information. For rest of the datasets, apart from these 7 classes, samples from the rest of the classes consider as unknown class and we consider samples from the whole dataset without any label information. Within the unknown class, some classes are domain specific which are not shared with the other domains.

\subsection{Model architecture and Training protocol:}
\noindent\textbf{MDCGAN:} We extended the algorithm proposed in \cite{FRIDA} by implementing both the feature generator $\mathcal{F}^{gan}$ and discriminator $\mathcal{M}^{gan}$ using multi-layer perceptrons consisting of three layers each, with Leaky ReLU activation functions. Additionally, a linear layer was added on top of $\mathcal{M}^{gan}$ for $\mathcal{B}^{gan}$, $\mathcal{C}^{gan}$, and $\mathcal{D}^{gan}$ individually. The latent variable $\textbf{z}$ was set to a dimensionality of 2000, and the input dimensionality of $\mathcal{F}^{gan}$ was defined as $(2000 + y^{dim} + d^{dim})$. Here, $y^{dim}$ represents the dimensionality of the one-hot encoding vector for closed classes, using a 3-bit binary encoding for each domain, and $d^{dim}$ has a dimensionality of 3. The output dimension of the feature generator $\mathcal{F}^{gan}$ was specified as 2048, which was fed as input to the discriminator $\mathcal{M}^{gan}$ with the same dimensionality of 2048.\\

\noindent\textbf{MEOSDA:}
In MEOSDA, the feature extractor $\mathcal{F}^{osda}$ is constructed using a multi-layer perceptron with three linear layers. The input dimension is 2048, followed by two fully connected layers of dimensions 1024 and 512, respectively. Depending on the number of additional domains, multiple classifier branches are added. Each domain (except $\mathcal{D}_1$) has a dedicated classifier branch comprising two fully connected layers with 256 and K + 1 nodes (where K is the number of known classes). Batch normalization and Leaky ReLU non-linearity are applied after each new layer to ensure stable training.

For $\mathcal{S}$ in MEOSDA we generate 100 samples per class of the previous domains from MDCGAN.For pseudo-labeling we fixed the probability threshold(th) as $0.95$. The MEOSDA network is trained using Adam optimizer with a learning rate set to 0.001, $\beta_1=0.5$, $\beta_2=0.9$, and a batch size of 64.

\subsection{Evaluation:} We compare our algorithm against methods like \cite{STA}, \cite{AOD}, and \cite{osdabp} on Office-Home, UPRN-RSDA and DomainNet datasets. While these are open-set domain adaptation methods, we enable their application in an incremental setting by using them in conjunction with incremental algorithms like \cite{EWC} and \cite{LWF}. Correspondingly, we use the exact same environment with the same open and closed classes for the evaluation of these methods as we use for our algorithm. The scarcity of the related methods to compare with our method we clubbed the Open set domain adaptation and Incremental learning methods.Tables \ref{tab:res1}, \ref{tab:res2} and \ref{tab:ref3} demonstrate the results of our methods and offer a direct comparisons with other open set domain adaptation methods in Incremental setup on  UPRN-RSDA, Office-Home and DomainNet, respectively.\\

\noindent \textbf{Evaluation protocols}: We report the average performances in terms of OS and OS* described in \cite{osdabp} over all the time stamp of each domain $\mathcal{D}_{\tau}$, $\tau \geq 2$ and defined it by $A$. Subsequently, the mean over all the domains are also mentioned. We report a measure of forgetting to highlight the evolution of the test performances for a given domain over the time. Precisely, for a domain $\mathcal{D}_{\tau}$, let $\mathcal{A}_{\tau}$ and $\mathcal{A}_{\tau+1}$  denote the performances at time $\tau$ and $\tau + 1$. Given that, the average forgetting for $\mathcal{D}_{\tau}$ till time $\tau$ is defined by $F(\mathcal{D}_{\tau})$ whereas the mean forgetting over all the domains is measured by averaging the $F(\mathcal{D}_{\tau})$ scores over all the domains for a given dataset. The mean forgetting over all domains is given by $ F(\mathcal{D}_{\tau}) = \frac{1}{T} \underset{k=1}{\overset{T}{\sum}} \mathcal{A}_{k+1} - \mathcal{A}_k$, T is number of increment. For a domain $\mathcal{D}_{m}$, $m\in \{ 2, ... , \tau \}$ at the time-stamp $\tau+1$, the number of increment T = $\tau - m$ .  A positive $(\mathcal{A}_{\tau+1} - \mathcal{A}_{\tau})$ indicates that the test accuracy increases at timestamp $\tau + 1$ than at timestamp $\tau$ for $\mathcal{D}_{\tau}$.


\begin{table*}
\centering
\scalebox{0.65}{
\begin{tabular}{|c|c|c|c|c|}
\hline
Method&\multicolumn{1}{|c|}{$\mathcal{D}_2$}  & \multicolumn{1}{|c|}{$\mathcal{D}_3$} & \multicolumn{1}{|c|}{$\mathcal{D}_4$} &
\multicolumn{1}{|c|}{AVG}\\
\hline

&~~~~~~~~OS~~~~~~~~~~~~~~~~OS*&~~~~~~~~OS~~~~~~~~~~~~~~~~OS*
&~~OS~~~~OS*
&~~~~~~~~OS~~~~~~~~~~~~~~~~OS*\\

&A~~~~~~~~~F~~~~~~~~~A~~~~~~~~~F
&A~~~~~~~~~F~~~~~~~~~A~~~~~~~~~F
&A~~~~~~~A
&A~~~~~~~~~F~~~~~~~A~~~~~~~~~F\\

\hline EWC(OSDA-BP)\cite{EWC},\cite{osdabp}
&48.26~~-24.93~~49.72~~-23.96
&60.75~~-11.54~~61.04~~-11.40
&12.93~~14.78
&40.65~~-18.23~~41.85~~-17.68\\
\hline

LWF(OSDA-BP)\cite{LWF},\cite{osdabp}
&47.36~~-32.71~~47.60~~-33.37
&55.83~~-23.14~~53.08~~-24.36
&6.04~~6.90
&36.41~~-27.93~~35.86~~-28.86\\
\hline

EWC(STA)\cite{EWC},\cite{STA}
&64.99~~-11.93~~67.01~~-11.14
&71.82~~0.47~~68.25~~0.45
&39.79~~44.18
&58.87~~-5.73~~59.81~~-5.35\\

\hline 

LWF(STA)\cite{LWF},\cite{STA}
&74.68~~-10.36~~74.93~~-9.25
&84.32~~-0.00~~82.17~~-0.09
&43.26~~48.20
&67.42~~-5.18~~68.43~~-4.67\\

\hline EWC(ATD)\cite{EWC},\cite{AOD}
&60.13~~-12.60~~60.94~~-10.80
&77.78~~-3.76~~75.98~~-3.29
&51.00~~51.36
&62.97~~-8.18~~62.76~~-7.04\\

\hline LWF(ATD)\cite{LWF},\cite{AOD}
&56.21~~-14.98~~56.98~~-14.12
&77.75~~-0.96~~75.38~~-1.37
&51.09~~49.35
&61.68~~-7.97~~60.57~~-7.74\\

\hline \textbf{Ours}
&\textbf{81.67~~-1.25~~80.63~~-1.10}
&\textbf{85.80~~-0.10~~83.95~~-0.45}
&\textbf{72.20~~70.70}
&\textbf{79.89~~-0.67~~78.43~~-0.77}\\
\hline

\end{tabular}
}
\caption{The performance comparison on the UPRN-RSDA dataset. We report the average performance (A) and forgetting (F) for each domain as well as the entire dataset. A negative average forgetting denotes that on an average, there is a performance drop for the domain. }
\label{tab:res1}
\end{table*}

\begin{table*}
\centering
\scalebox{0.68}{
\begin{tabular}{|c|c|c|c|c|}
\hline
Method&\multicolumn{1}{|c|}{$\mathcal{D}_2$}  & \multicolumn{1}{|c|}{$\mathcal{D}_3$} & \multicolumn{1}{|c|}{$\mathcal{D}_4$} &
\multicolumn{1}{|c|}{AVG}\\
\hline

&~~~~~~~~OS~~~~~~~~~~~~~~~~OS*
&~~~~~~~~OS~~~~~~~~~~~~~~~~OS*
&~~OS~~~~OS*
&~~~~~~~~OS~~~~~~~~~~~~~~~~OS*\\

&A~~~~~~~~~F~~~~~~~~~A~~~~~~~~~F
&A~~~~~~~~~F~~~~~~~~~A~~~~~~~~~F
&A~~~~~~~A
&A~~~~~~~~~F~~~~~~~A~~~~~~~~~F\\

\hline EWC(OSDA-BP)\cite{EWC},\cite{osdabp}
&44.40~~-26.02~~44.16~~-25.57
&21.45~~-1.23~~21.48~~-1.27
&7.14~~7.32
&24.33~~-13.63~~24.32~~-13.42\\
\hline LWF(OSDA-BP)\cite{LWF},\cite{osdabp}
&43.08~~-24.49~~42.70~~-24.03
&21.63~~-0.13~~22.17~~-0.14
&9.52~~9.76
&24.74~~-12.31~~24.88~~-12.08\\
\hline EWC(STA)\cite{EWC},\cite{STA}
&61.23~~-24.13~~60.99~~-23.64
&35.62~~-1.05~~35.73~~-1.03
&18.32~~18.77
&38.39~~-12.59~~38.50~~-12.34\\
\hline LWF(STA)\cite{LWF},\cite{STA}
&59.55~~-25.91~~59.32~~-25.38
&33.90~~-0.06~~33.74~~-0.06
&11.09~~11.37
&34.85~~-12.98~~34.81~~-12.72\\

\hline EWC(ATD)\cite{EWC},\cite{AOD}
&47.22~~-26.60~~47.05~~-26.17
&24.76~~-2.53~~25.25~~-2.46
&6.50~~6.66
&26.16~~-14.57~~26.32~~-14.32\\
\hline LWF(ATD)\cite{LWF},\cite{AOD}
&47.83~~-26.91~~47.51~~-26.40
&23.05~~-2.61~~23.51~~-2.54
&7.53~~7.72
&26.14~~-14.76~~26.25~~-14.47\\

\hline \textbf{Ours}
&\textbf{76.25~~0.41~~75.58~~0.11}
&\textbf{55.78~~0.03~~54.52~~0.92}
&\textbf{63.47~~61.71}
&\textbf{65.16~~0.22~~63.94~~0.51}\\
\hline
\end{tabular}
}
\caption{The performance comparison on the Office Home dataset. We report the average performance (A) and forgetting (F) for each domain as well as the entire dataset. A negative average forgetting denotes that on an average, there is a performance drop for the domain.} \label{tab:res2}
\end{table*}

\begin{figure*}[ht]
    \centering
    \begin{minipage}[b]{0.47\linewidth}
    \fbox{\includegraphics[width=0.95\linewidth, height=4cm]{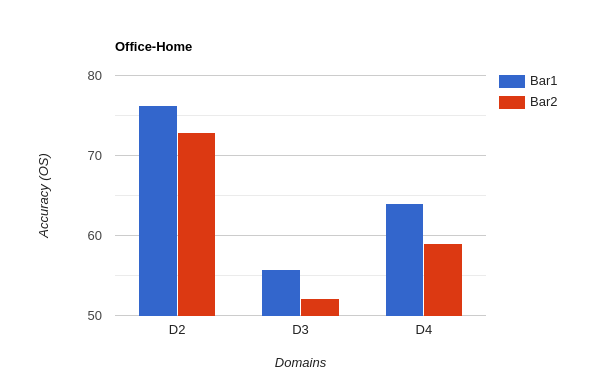}}
    \caption{Office home }
    \label{fig:ablation2_multihead_offH}
    \end{minipage}
 \quad
 \begin{minipage}[b]{0.47\linewidth}
    \fbox{\includegraphics[width=0.97\linewidth, height=3.97cm]{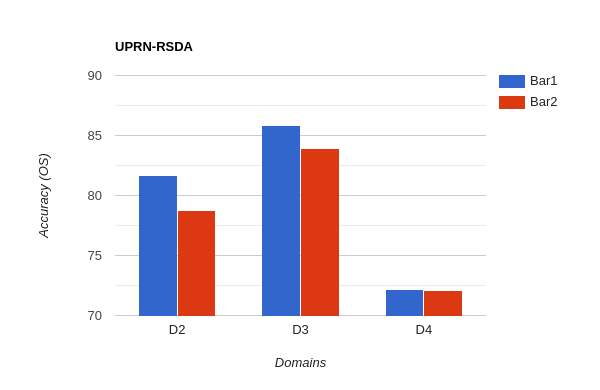}}
    \caption{RS  $~~~~~~~~~~~~~~~~~~~~~~~~~ $                                                  }
    \label{fig:ablation2_multihead_RS}
    \end{minipage}
    \caption{effect of multi head classifier compared to single head classifier}
\end{figure*}

\begin{table*}
\centering
\scalebox{0.5}{
\begin{tabular}{|c|c|c|c|c|c|c|}
\hline
Method&$\mathcal{D}_2$  & $\mathcal{D}_3$ & $\mathcal{D}_4$ & $\mathcal{D}_5$ & $\mathcal{D}_6$ & AVG \\
\hline
&OS~~~~~~~~~~~~OS* & OS~~~~~~~~~~~~OS* & OS~~~~~~~~~~~~OS* & OS~~~~~~~~~~~~OS* & OS~~~~~~~~~~~~OS* & OS~~~~~~~~~~~~OS*\\
&A~~~~~~F~~~~~~A~~~~~~F & A~~~~~~F~~~~~~A~~~~~~F & A~~~~~~F~~~~~~A~~~~~~F & A~~~~~~F~~~~~~A~~~~~~F & A~~~~~~A & A ~~~~~~F~~~~~~A~~~~~~F\\

\hline EWC(OSDA-BP)\cite{EWC},\cite{osdabp}
&7.92~~-5.13~~7.91~~-5.12
&3.04~~-2.25~~3.05~~-2.26
&0.35~~-0.49~~0.35~~-0.49
&0.19~~-0.10~~0.19~~-0.10
&0.09~~0.09
&2.32~~-1.99~~2.32~~-1.99\\
\hline LWF(OSDA-BP)\cite{LWF},\cite{osdabp}
&7.92~~-5.22~~7.89~~-5.21
&3.03~~-2.13~~3.04~~-2.14
&0.54~~-0.53~~0.55~~-0.54
&0.21~~-0.25~~0.21~~-0.25
&0.04~~0.04
&2.35~~-2.04~~2.34~~-2.03\\
\hline EWC(STA)\cite{EWC},\cite{STA}
&13.45~~-8.09~~13.44~~-8.08
&7.85~~-5.47~~7.88~~-5.49
&2.08~~-1.84~~2.09~~-1.85
&0.64~~-0.09~~0.64~~-0.08
&0.63~~0.63
&4.93~~-3.87~~4.94~~-3.88\\
\hline LWF(STA)\cite{LWF},\cite{STA}
&13.16~~-8.15~~13.14~~-8.14
&7.29~~-5.10~~7.32~~-5.12
&1.94~~-1.71~~1.95~~-1.72
&0.45~~-0.13~~0.45~~-0.13
&0.40~~0.40
&4.65~~-3.77~~4.65~~-3.77\\
\hline EWC(ATD)\cite{EWC},\cite{AOD}
&8.25~~-6.18~~8.22~~-6.16
&2.26~~-1.68~~2.27~~-1.68
&0.41~~-0.46~~0.41~~-0.46
&0.07~~-0.08~~0.07~~-0.08
&0.01~~0.01
&2.20~~-2.10~~2.20~~-2.10\\
\hline LWF(ATD)\cite{LWF},\cite{AOD}
&8.55~~-6.35~~8.52~~-6.34
&2.55~~-1.89~~2.56~~-1.90
&0.60~~-0.62~~0.60~~-0.62
&0.07~~-0.10~~0.08~~-0.10
&0.02~~0.02
&2.36~~-2.24~~2.36~~-2.24\\

\hline \textbf{Ours}
&25.18~~-2.61~~24.30~~-2.82
&25.88~~-0.72~~21.82~~-0.65
&15.83~~-0.55~~14.98~~-0.55
&12.36~~-1.09~~12.71~~-0.04
&1.24~~0.93
&16.10~~-1.24~~14.95~~-1.02\\
\hline
\end{tabular}
}
\caption{The performance comparison of OS on the DomainNet dataset. We report the average performance (A) and forgetting (F) for each domain as well as the entire dataset. A negative average forgetting denotes that on an average, there is a performance drop for the domain. }\label{tab:ref3}
\end{table*}

\section{Analysis}
\subsection{Effect of using multi-head classifier}
From the bar graph visualization it is clearly seen that the effect of multi-head classifier where each classifier head is specified for a particular source domain at that time-stamp compare to using a single classifier for all the domains. The bar-graph visualization on Office-Home (Fig-\ref{fig:ablation2_multihead_offH})  and UPRN-RSDA  (Fig-\ref{fig:ablation2_multihead_RS}) dataset showing us the significant improvement compare to single head classifier. Bar1 is depicting results on multi-head classifier and Bar2 depicting as single head classifier.

\subsection{Domain Alignment visualisation}
\begin{figure*}[ht]
    \centering
    \begin{minipage}[b]{0.2\linewidth}
    \fbox{\includegraphics[width=0.97\linewidth, height=3cm]{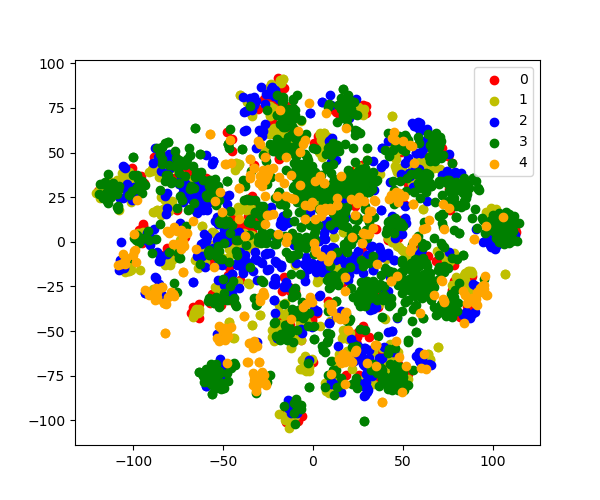}}
    \caption{  }
    \label{fig:adaptation_bad}
    \end{minipage}
 \quad
 \centering
    \begin{minipage}[b]{0.2\linewidth}
    \fbox{\includegraphics[width=0.97\linewidth, height=3cm]{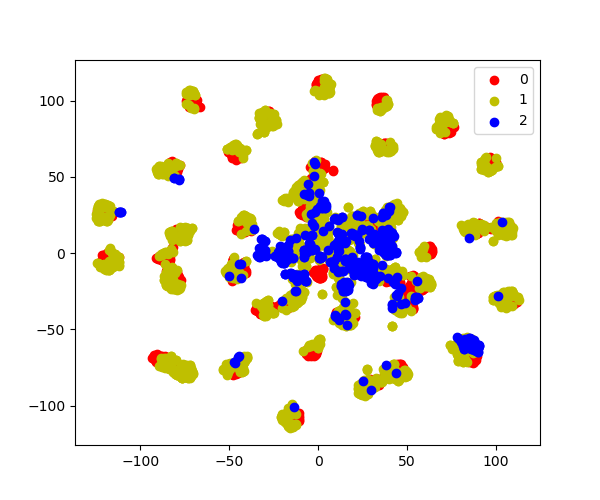}}
    \caption{ }
    \label{fig:adaptation_1st}
    \end{minipage}
 \quad
 \centering
    \begin{minipage}[b]{0.2\linewidth}
    \fbox{\includegraphics[width=0.97\linewidth, height=3cm]{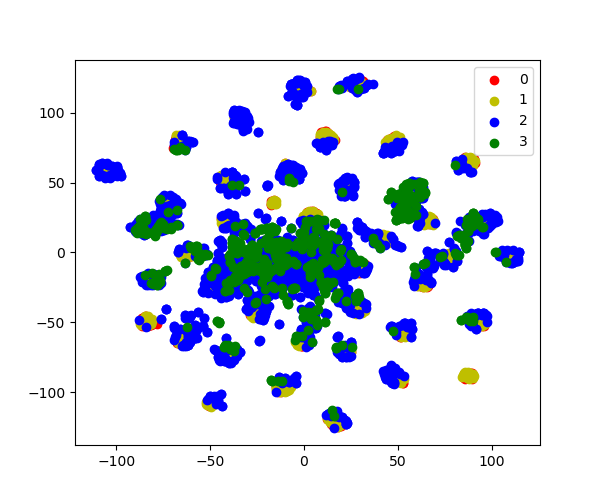}}
    \caption{ }
    \label{fig:adaptation_2nd}
    \end{minipage}
 \quad
 \begin{minipage}[b]{0.2\linewidth}
    \fbox{\includegraphics[width=0.97\linewidth, height=3cm]{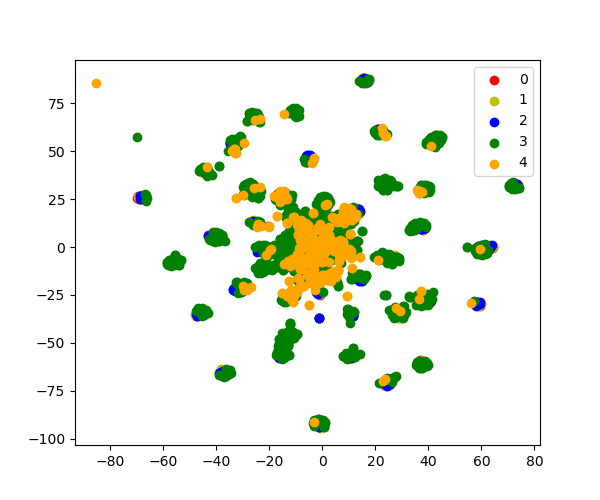}}
    \caption{  }
    \label{fig:adaptation_all}
    \end{minipage}
    \caption{Domain alignment visualization of Office-Home datadet, At, Fig-\ref{fig:adaptation_bad} depict t-SNE plot without domain adoptation where 0, 1, 2, 3 indicates domains $\mathcal{D}_{1}$, $\mathcal{D}_{2}$, $\mathcal{D}_{3}$, $\mathcal{D}_{4}$ respectively and 4 depicts as open-class samples from all the domains except $\mathcal{D}_{1}$ . At, timestamp $\tau = 1$, Fig-\ref{fig:adaptation_1st}, 0, 1 indicates domains $\mathcal{D}_{1}$, $\mathcal{D}_{2}$ respectively and 2 depicts as open-class samples. At, timestamp $\tau = 2$, Fig-\ref{fig:adaptation_2nd}, 0, 1, 2 indicates domains $\mathcal{D}_{1}$, $\mathcal{D}_{2}$, $\mathcal{D}_{3}$ respectively and 3 depicts as open-class samples. and At, timestamp $\tau = 2$, Fig-\ref{fig:adaptation_all}, 0, 1, 2, 3 indicates domains $\mathcal{D}_{1}$, $\mathcal{D}_{2}$, $\mathcal{D}_{3}$, $\mathcal{D}_{4}$ respectively and 4 depicts as open-class samples.}
\end{figure*}

The t-SNE generation on different domains before and after domain adaptation at different time-stamp showing us the domain alignment capabilities of MEOSDA on closed-set labels and at the same time segregation of open-set class labels. From the picture it is clear that our model performed well on the given task of classifying closed set classes and segregating open-set classes for the closed set classes.   Fig- \ref{fig:adaptation_bad}, Fig-\ref{fig:adaptation_1st}, Fig-\ref{fig:adaptation_2nd} and Fig-\ref{fig:adaptation_all} shows the visualization before adaptation, adaptation at timestamp $\tau = 1$, adaptation at timestamp $\tau = 2$, adaptation at timestamp $\tau = 3$, respectively.

\section{Conclusion}
In summary, we have introduced a novel problem of incremental open-set domain adaptation and proposed an effective solution. To tackle this challenge, we employ an ensemble classification technique to evaluate the intricacies of the multi-domain classifier. Our extensive experiments on diverse benchmark datasets have demonstrated the effectiveness of our model. The newly developed MDCGAN is not only beneficial for domain generation in incremental learning but also applicable in scenarios requiring multi-domain and multi-class sample generation. The results from various ablation studies further confirm the efficacy of our proposed models, although there is potential for improvement in refining the problem definition and corresponding model adjustments. Future work could explore scenarios where open classes are initially present in the source domain from the outset.

%
%
\bibliographystyle{splncs04}
\bibliography{main}
\end{document}